\documentclass[11pt,a4paper,fleqn]{article}
\usepackage[hyperref]{acl2020}
\usepackage{times}
\usepackage{latexsym}

\usepackage{lscape}
\usepackage{tikz}
\usepackage{graphicx}
\usepackage{leftidx}
\usepackage{tikz-cd}
\usepackage{listings}
\usepackage{mathtools}
\usepackage{blkarray, bigstrut}
\usepackage{tikz-dependency}
\usepackage{fancyhdr}
\usepackage[utf8]{inputenc}
\usepackage{amssymb}
\usepackage{latexsym}
\usepackage{enumerate} 
\usepackage{amsthm}
\usepackage{qtree}

\usepackage{microtype}

\title{Scalable knowledge base completion with superposition memories}
\author{Matthias Lalisse\\
Dept of Cognitive Science\\
Johns Hopkins University\\
Baltimore, MD USA \\
{\tt lalisse@jhu.edu} \And
Eric Rosen\\
Dept of Cognitive Science\\
Johns Hopkins University\\
Baltimore, MD USA \\
{\tt erosen27@jhu.edu} \And
Paul Smolensky\\
Dept of Cognitive Science\\
Johns Hopkins University\\
\& Microsoft Research AI \\
Redmond, WA USA\\
{\tt smolensky@jhu.edu}
}

\date{\today}

\aclfinalcopy

\begin{document}

\maketitle

\begin{abstract}
We present Harmonic Memory Networks (HMem), a neural architecture for knowledge base completion that models entities as weighted sums of pairwise bindings between an entity's neighbors and corresponding relations. Since entities are modeled as aggregated neighborhoods, representations of unseen entities can be generated on the fly. We demonstrate this with two new datasets: WNGen and FBGen. Experiments show that the model is SOTA on benchmarks, and flexible enough to evolve without retraining as the knowledge graph grows.
\end{abstract}

\section{Introduction} 

The existence of large but inexhaustive databases of specialized (e.g.\ WordNet) and general world (e.g.\ Freebase) knowledge has motivated the development of methods that allow such databases to be automatically extended using computational methods|knowledge base completion (KBC). 
Typical approaches employ an embedding-based strategy: elements of a fact|which in this task setting come in the form of triplets consist of a pair of entities and a relation, e.g.\ $(\mathtt{spy\_{}kids}, \mathtt{has\_{}actor}, \mathtt{mike\_{}judge})$|are combined into a representation of the fact via some systematic function, and then scored. 
Methods of this sort generally rely on learned embeddings of fact elements into low-dimensional vector spaces. 
Since representations must be learned from training-set instances of each component, this creates problems when such databases are to be scaled, and therefore these methods have difficulty accommodating an open-world setting in which knowledge graphs evolve in time, since new facts inserted into the database after model training cannot be used for inference without model retraining. Furthermore, databases may be augmented in time not only with new facts about known entities, but also with new entities. In embedding-based models, new representation for such entities must be trained.

We present Harmonic Memories (HMem), a neural network which models entities by aggregating information about their neighborhoods using a superposition memory architecture, achieving generalization to new entities without retraining.\footnote{Code and datasets are available at \url{github.com/MatthiasRLalisse/HMemNetworks}.} The network combines two ideas. First, a representation of entities as memory states consisting of superposed vector associations between learned entity and relation embeddings. Second, completion of memory states using a learned transformation based on Harmony-optimization methods \cite{smolensky_legendre2006harmonic_mind} (see \S\ref{sec_optimization}). We refer to vector associations as \emph{bindings} in the sense of the "variable-binding problem" in the philosophy of cognitive science: in neural net models of cognition, how are representations of the elements of a structure bound together into structures? In this work, we investigate two solutions prominent in the cognitive science literature|tensor product binding \citep{smolensky1990tensor} and circular convolution \citep{plate1994thesis}|which have also both been effectively applied in KBC \citep{nickel2011rescal,nickel2016hole}.

The approach is inspired by computational modeling of biological neural architectures for knowledge representation \citep{crawford}, and is related to KBC methods based on convolution of graph neighborhoods \cite{schlichtkrull2017graphconv, dettmers2018convE, nguyen2018convembed}, in which inference is performed over representations of aggregated entity neighborhoods. Recent work has extended this idea using Graph Attention Networks (GANs) \citep{nathani2019GraphAttention}, which assign attention weights to entries in a graph neighborhood, these being later combined. For instance, \citet{velickovic2018GraphAttention} use Graph Attention to generate weights for triplet representations obtained by transforming concatenated entity and relation vectors, combining the results by averaging. This is similar to our approach, with the key difference that formulating the model|as we do|in terms of binding allows for clear formal analysis of certain scaling results (\S\ref{section_saturation_effects}). We therefore gain in interpretability.

HMem scales well in three respects. First, it allows a database with a fixed set of entities and relations to incorporate new facts into the model without parameter re-estimation. Empirically, performance improves in nearly every case when the neighborhoods are thus expanded. Second, it permits the addition of entities unseen in training, whose representations are useless in a vector embedding framework. For our model, inferences about these entities are possible when a subgraph including them becomes available. Third, our model effectively handles nodes with high in-degree. We show that, whereas embedding-based approaches show decreased performance with highly connected nodes, our model exhibits improved performance on nodes with many neighbors.

\S\ref{section_representing_memory_states}, \S\ref{section_binding_models} and \S\ref{sec_optimization} introduce the Harmonic Memory architecture, and \S\ref{section_results} shows that our model achieves state-of-the-art results on benchmark KBC datasets. After evaluation on standard benchmarks, \S\ref{section_generalizing} introduces WNGen and FBGen, datasets based on WordNet and \textsc{Freebase} that evaluate the network's ability to abstract from node identity and make inferences exclusively on the basis of information about nodes in its neighborhood, and \S\ref{section_saturation_effects} examines in detail how the model scales with the size of entity neighborhoods and the addition of new input facts. \S\ref{section_conclusion} concludes.

\section{Representation of memory states} \label{section_representing_memory_states}

A knowledge graph consists of triplets composed of a pair of entities and a relation, e.g. $\left(\boldsymbol{e}_\text{cat}, \boldsymbol{r}_\text{has\_{}part}, \boldsymbol{e}_\text{paw}\right)$. Here, we say that $cat$ is a left-neighbor of $paw
$ with respect to the relation $has\_{}part$. In graph completion, we are given a query in the form of either $\left(\cdot, \boldsymbol{r}, \boldsymbol{e}_r\right)$ or $\left( \boldsymbol{e}_\ell, \boldsymbol{r}, \cdot\right)$, representing queries of the left and right entity respectively. 

We denote discrete symbols with calligraphic fonts (e.g.\ $\mathcal{V}$), corresponding vector spaces with italics (e.g.\ $V$), individual symbols from $\mathcal{V}$ with bold letters, corresponding vectors in $V$ in italics (e.g.\ $\boldsymbol{v}_i$ is a symbol and $v_i$ its vector embedding) and memory states with $\mathtt{M}$.  $\mathcal{E}$ and $\mathcal{R}$ denote sets of entities and relations respectively. We embed each entity and relation symbol $\boldsymbol{e}_i \in \mathcal{E}, \boldsymbol{r}_i \in \mathcal{R}$ in a $d_e, d_r$-dimensional space, yielding vectors $e_i\in E$ and $r_i\in R$ which are used to construct graph memories $\mathtt{M}_i$ for each entity by aggregating triplet entries (see Section \ref{section_binding_models}) into a representation of the entity's immediate neighborhood \ref{section_binding_models}. For example, the memory state for $e_{canine}$ would include bindings of \emph{hypernym} to \emph{dog}, \emph{has\_{}part} to \emph{paw}, \emph{hyponym} to \emph{mammal}, etc.
%
\begin{figure}[!ht]
\centering 
\includegraphics[scale=0.18]{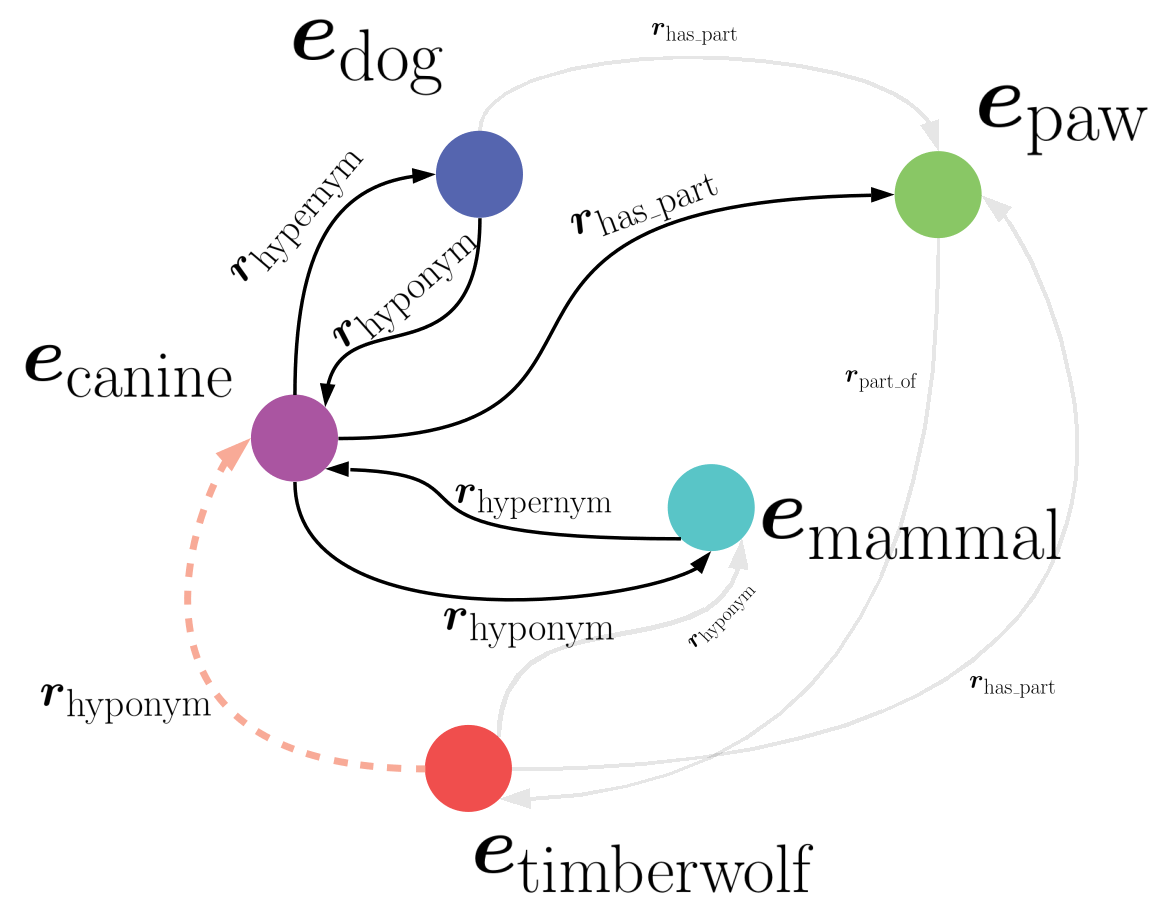}
\caption{Neighborhood subgraph for an entity $\boldsymbol{e}_\text{canine}$. In this example, the memory state $\mathtt{M}_\text{canine}$ in the query $(\cdot,\boldsymbol{r}_\text{hyponym}, \boldsymbol{e}_\text{canine})$ with true completion $\boldsymbol{e}_\text{timberwolf}$ is 
$\mathbb{B}(r^r_{\text{hyper}},e_\text{dog})) + 
    \mathbb{B}(r^\ell_{\text{hypo}},e_\text{dog} ) +
    \mathbb{B}(r^r_{\text{has\_part}},e_\text{paw}) + 
    \mathbb{B}(r^\ell_{\text{hypo}},e_\text{mammal}) + 
    \mathbb{B}(r^r_{\text{hyper}}, e_\text{mammal})$.
} \label{fig_neighborhood_example}
\end{figure}

Each model in the HMem class is parametrized by a binding map $\mathbb{B}$ that associates entity and relation entries, and a corresponding unbinding map $\mathbb{U}$. 
Unbinding approximately inverts binding operation by addressing the memory using a query relation vector ${r}_q$ to retrieve entity vectors likely to be bound with that relation in the memory. The binding and unbinding maps are chosen such that, for an entity whose neighborhood is the singleton set $\{(\boldsymbol{e}_i,\boldsymbol{r}_j)\}$:
\begin{align*}
\mathbb{U}\left(r_j, \mathbb{B}\left(\{(e_i,r_j)\}\right)\right) \approx e_i
\end{align*}
i.e., $e_i$ can be approximately retrieved from the binding of $e_i$ to $r_j$ by addressing the memory $\mathtt{M} = \mathbb{B}\left(\{(r_j,e_i)\}\right)$.

Associations between entity and relation vectors in our model are pairwise. To account for the directionality of relations, we train two embeddings for each relation|one associated with an entity's left-neighbors and another for its right-neighbors. Each relation thus has a left-embedding $r^\ell$ and a right-embedding $r^r$, with, for instance, entries $(\boldsymbol{r}^r_\text{has\_part}, \boldsymbol{e}_\text{claw})$ in the memory state for \emph{feline} and an entry $(\boldsymbol{r}^\ell_\text{has\_part}, \boldsymbol{e}_\text{feline})$ in the memory state for \emph{claw} (see Fig. \ref{fig_neighborhood_example}).

\begin{figure*}[!ht]
\centering
\includegraphics[scale=.18]{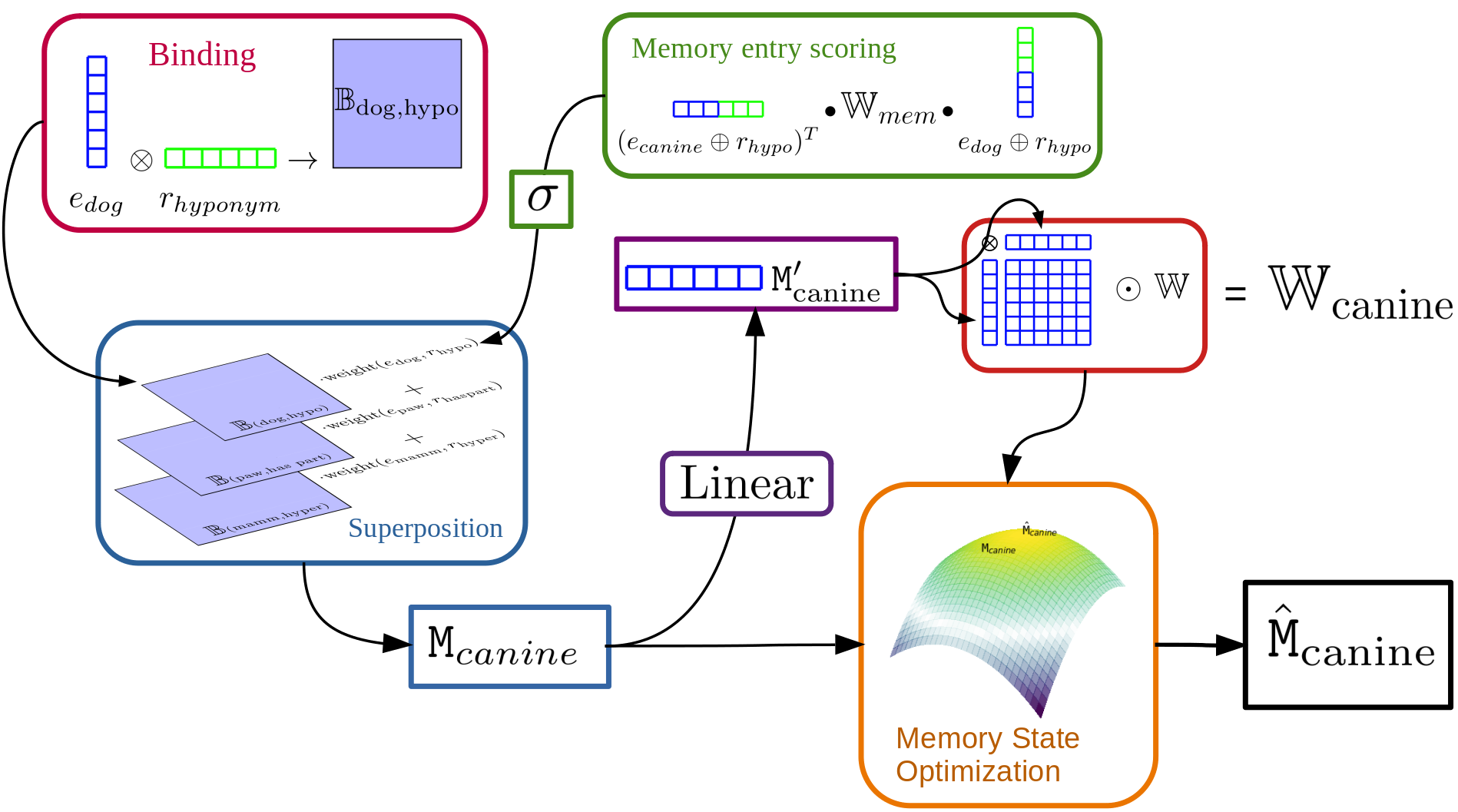}
\caption{Harmonic Memory architecture with TPR binding.}
\end{figure*}

\section{Binding} \label{section_binding_models}

Harmlessly overloading the symbol $\mathbb{B}$ to apply to both tuples and sets of tuples, we consider binding operations that aggregate neighborhoods via summation of the individual entries, i.e.
\begin{align*}
\mathbb{B}\left(\{(r_i, e_i)\}\right) = \sum_{i} \mathbb{B}\left(r_i,e_i\right)
\end{align*}
where $\mathbb{B}\left(r_i, e_i\right)$ is the binding of the i$^{th}$ memory entry. As binding operations, we evaluate the tensor product (\ref{def_tpr_binding}) \cite{smolensky1990tensor} and  circular convolution (\ref{def_ccorr_binding}). 
\begin{align}
\mathbb{B}^\text{TPR}(r, e) = r \otimes e \label{def_tpr_binding} \\
\mathbb{B}^\text{CConv}(r,e) = r\circledast e \label{def_ccorr_binding}
\end{align}
A tensor product is unbound by left-dotting the memory state with a relation vector $r$: 
\begin{equation}
\mathbb{U}^\text{TPR} \left(r, \mathtt{M}\right)  = r \cdot \mathtt{M}
\end{equation}
When the relation vectors are normalized, $\mathbb{U}^\text{TPR}$ exactly recovers $e$ from the singleton memory state $\mathtt{M} = r \otimes e$. For CConv, the unbinding operation is
\begin{equation}
\mathbb{U}^\text{CConv}\left(r, \mathtt{M}\right) = r \star \mathtt{M}
\end{equation}
$\star$ denotes circular correlation, which is computed efficiently using the Fourier transform $r \star \mathtt{M} = \mathcal{F}^{-1}\left(\overline{\mathcal{F}(r)} \odot \mathcal{F}\right)$, $\overline{x}$ denoting the complex conjugate of $x$. 

Circular convolution was introduced to connectionist modeling by \citet{plate1994thesis}, and applied to KBC by \citet{nickel2016hole} and \citet{lalisse2019gradientgraphs}, exploiting the fact that the correlation is an approximate inverse of convolution ($x \star (x \circledast y)  \approx y$), under additional stipulations discussed in Appendix 1. There, we describe a transformation on embeddings in the CConv model that guarantee this property, and improved performance on the CConv models.

\textbf{Memory weighting.} For entities with large neighborhoods, it is intractable to compute bindings for all neighbors.\footnote{The largest entity neighborhood in WordNet contains 961 links (mean=7), and 9739 (mean=65) for Freebase.} So, prior to superposition, we filter candidate bindings and commit them to memory in graded form. Entity and relation pairs are scored with respect to the $e_i$ and the query relation $r_q$ according to Eqn. (\ref{eqn_binding_ranks}), where $\oplus$ denotes vector concatenation. $W_\text{weight}$ and $b_\text{weight}$ are learned weight matrices and bias vectors indexed to $r_q$. 
\begin{align}
\begin{split}
 \scriptstyle \text{weight}(e_c, r_c|e_i, r_q) = 
    &  \scriptstyle \small \sigma { (} \left(e_i \oplus r_q\right)^\top W_\text{score}\left(e_c \oplus r_c\right) \\
    & \scriptstyle + {b_{score}^q}^\top \left(e_c \oplus r_c \right) { )} 
\end{split} \label{eqn_binding_ranks}
\end{align}
After scoring, the top $k=200$ candidate neighbors are bound and entered into memory, weighted by their scores:\footnote{While the memory state $\mathtt{M}_i$ for a given query also depends on the entity query relation $r_q$, this additional subscript is omitted for convenience. }
\begin{align}
\mathtt{M}_i = \sum_{c} \text{weight}(r_c,e_c|e_i, r_q) \mathbb{B}\left(r_c,e_c\right) 
\end{align} \label{eqn_score_weighted_binding}

\textbf{Remark.} Our binding-based approach is inspired by \citet{crawford}, who developed a biologically realistic neural network for representing WordNet. 
In their model, a memory state vector for each entity is formed by summing pairwise associations (convolution) of entities and relations, one association per graph link. Links can be recovered by unbinding stored associations from entity memory states to recover a node's immediate neighbors. Since Crawford et. al. are mainly preoccupied with neural realism rather than learning or generalization, the embeddings for each graph element are untrained, and they evaluate their model on an embedding of the full WordNet database in a simple artificial task (graph traversal). Thus, their work only investigates the model's ability to robustly retrieve the vectorized knowledge graph. Since our target task requires generalization from a partial graph, we introduce additional operations that complete the representation of each entity.

\section{Memory completion} \label{sec_optimization}

The assembled memory state is forwarded to a memory completion operation based on optimization of the Harmony Equation (\ref{eqn_harmony})|which is parametrized by a learned symmetric weight matrix $\mathbb{W}$ and bias vector $b$|with respect to the vector $m$.
\begin{align}
\begin{split}
\mathcal{H}_{\mathbb{W},b}(\mathtt{M}_i, m) = &\frac{1}{2}\left( m^\top \mathbb{W} m + b^\top \right)\\ & - \frac{\lambda}{2} \left(\mathtt{M}_i - m\right)^\top \left(\mathtt{M}_i - m\right)\end{split} \label{eqn_harmony}
\end{align}
This is solved by
\begin{align}
    \begin{split}
    \hat{\mathtt{M}}_i 
    & = \text{argmax}_m \mathcal{H}_{\mathbb{W},b}(\mathtt{M}_i, m) \\
    & = (\mathbb{W}_i - \lambda I)^{-1} ( 2\lambda \mathtt{M}_i + b) 
    \end{split} \label{eqn_max_mem_state}
\end{align}
when $\lambda$|a hyperparameter|is greater than the spectral norm of $\mathbb{W}$, guaranteeing the existence of a unique optimum for $\mathcal{H}_{\mathbb{W},b}$. This formulation is motivated by associative memory models like the formally similar Hopfield networks \citep{hopfield1982}, which complete corrupted input patterns by minimizing the Energy of the resulting network configuration. 
$\lambda$ controls the magnitude of a penalty for the squared distance between the output and the input memory $\mathtt{M}_i$, with a value of $\lambda= \infty$ implying that $\mathcal{H}_{\mathbb{W},b}$ is maximized at $\mathtt{M}_i$ (the memory state remains where it is).

\begin{figure}
\centering
\includegraphics[scale=.25]{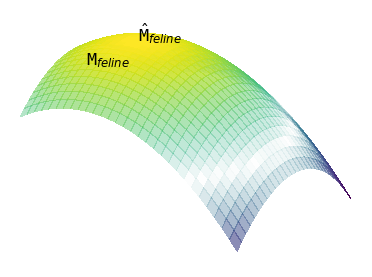}
\includegraphics[scale=.25]{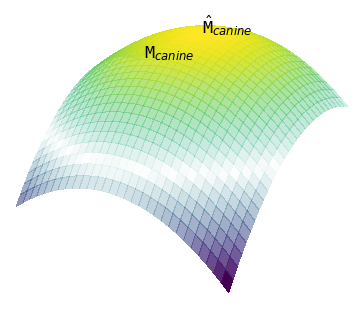}
\caption{Harmony functionals on a two-dimensional memory state space (visualized as surfaces) and optima defined by weight matrices $\mathbb{W}_\text{feline},\mathbb{W}_\text{canine}$ calculated from distinct memory states $\mathtt{M}_\text{feline},\mathtt{M}_\text{canine}$.} \label{fig_harmony_surfaces}
\end{figure}

We allow the parameters of the Harmony function to change with the query being posed by specifying a weight matrix $\mathbb{W}_i$ computed for any given $\mathtt{M}_i$. The local weight matrix $\mathbb{W}_i$ is computed from a global weight matrix $\mathbb{W}_\text{global}$ and a filter vector $M'_i$, which is a function of the input memory:
\begin{align}
\mathbb{W}_i &= {M_i'}{M_i'}^\top \odot \mathbb{W}_\text{global} \label{def_r_metric}
\end{align}
where 
\begin{align}
{M}_i' &= W_{map}\ \mathtt{M}_i + b_\text{map}
\end{align}
$W_\text{map}$ and $b_\text{map}$ are a learned matrix and bias vector mapping each memory state to a filter vector, whose self-outer product multiplies the global weight matrix elementwise. The resulting weight matrices vary smoothly with the value of the input memory state, leading to distinct hypersurfaces in the $|\mathtt{M}|$-dimensional space of memory states (e.g.\  figure \ref{fig_harmony_surfaces}.) 

In inference, we first optimize $\mathtt{M}_i$ to generate $\text{opt}(\mathtt{M}_i) = \hat{\mathtt{M}}_i$. The optimized memory state is then probed using the unbinding map $\mathbb{U} (r_q,\text{opt}(\mathtt{M}_i))$, 
returning a vector $e_o$ representing the output of a probe of the memory $\mathtt{M}_i$ of $\boldsymbol{e}_i$ for entities that are related to $\boldsymbol{e}_i$ via $\boldsymbol{r}_q$. The output of unbinding is then compared with the vectors $e_c$ for all candidate completions $\boldsymbol{e}_c$ using the squared Euclidean distance.
\begin{align*}
    \text{score}(\boldsymbol{e}_c) = \left\lVert e_o - e_c \right\rVert^2
\end{align*}


\textbf{Training.} During training, the link for the current training instance is withheld from the neighborhood for the corresponding entity, with the input memory state constructed from the remaining neighbors. This incomplete memory state is then optimized with respect to Eqn. (\ref{eqn_harmony}), and the result probed for a predicted completion. We use the cross entropy training loss derived from the squared Euclidean distance of the output from the true completion, relative to a negative sample $\mathcal{N} = \{ \boldsymbol{e}_n \}$ of alternative completions. For instance, the loss for a right-probe of $(\boldsymbol{e}_i,\boldsymbol{r},\cdot)$ with true entity $\boldsymbol{e}_j$ is
\begin{align*}
    \mathcal{L}(\boldsymbol{e}_j|\boldsymbol{e}_i,\boldsymbol{r}) = -\log \frac{\exp\{\left\lVert e_o- e_j \right\rVert^2\}}{\sum_{\boldsymbol{e}_n \in \mathcal{N}} \exp\{ \left\lVert e_o - e_n \right\rVert^2 \} }
\end{align*}

\begin{table*}
\centering 
{\scriptsize
\hspace{-.65cm}
\begin{tabular}{|l|ccccc|ccccc|}
\multicolumn{1}{c}{\phantom{h}}& \multicolumn{5}{c}{WordNet} & \multicolumn{5}{c}{Freebase} \\ \hline
Model & MR & MRR & H@1 & H@3 & H@10 & MR & MRR & H@1 & H@3 & H@10 \\ \hline\hline
DistMult \cite{yang2015distmult}$^\dagger$ & 457 & .790 & - & - & .950 & 36 & .837 & - & - & \textbf{\underline{.904}} \\
ComplEx \cite{troullion2016complex} & - & .941 & .936 & .945 & .947 & - & .692 & .599 & .759 & .840 \\
R-GCN+ \cite{schlichtkrull2017graphconv} & - & .819 & .697 & .929 & \textbf{\underline{.964}} & - & .696 & .601 & .760 & .842 \\
ConvE \cite{dettmers2018convE} & 374 & .943 & .935 & .946 & .956 & 51 & .657 & .558 & .723 & .831 \\ 
SimplE \cite{kazemi2018simplE} & - & .942 & .939 & .944 & .947 & - & .727 & .660 & .773 & .838 \\
HypER \cite{balazevic2019hyper} & \textbf{431} & \textbf{\underline{.951}} & \textbf{\underline{.947}} & \textbf{\underline{.955}} & .958 & 44 & \textbf{\underline{.790}} & \textbf{\underline{.734}} & \textbf{.829} & .885\\
TorusE \cite{ebisu2018torusE} & - & .947 & .943 & .950 & .954 & - & .733 & .674 & .771 & .832\\ 
\hline
HMem-CConv  & 262 & .927 & .913 & .939 & .946 & \textbf{\underline{24}} & .664 & .548 & .749 & .867 	\\ 
HMem-CConv+ & 227 & .933 & .919 & \textbf{.945} & \textbf{.952} & \textbf{\underline{24}} & .664 & .547 & .749 & .866 \\ 
HMem-CConv$_\infty$  & 308 & .884 & .851 & .912 & .934 & 39 & .488 & .363 & .554 & .734 \\
HMem-CConv$_\infty$+ & \textbf{183} & .899 & .866 & .930 & .951 & 39 & .481 & .357 & .546 & .725\\
HMem-CConv$_\text{im}$ & 344 & \textbf{.936} & \textbf{.929} & .942 & .947 & 25 & \textbf{.728} & \textbf{.637} & \textbf{.795} & \textbf{.881} \\
\hline
HMem-TPR & 253 & .934 & .923 & .944 & .948 & 30 & .590 & .478 & .660 & 788 \\ 
HMem-TPR+ & \textbf{\underline{174}} & \textbf{.944} & \textbf{.932} & \textbf{\underline{.955}} & \textbf{.960} & 29 & .592 & .479 & .662 & .791 \\
HMem-TPR$_\infty$ & 395 & .874 & .823 & .922 & .939 & 38 & .612 & .517 & .669 & .782 \\
HMem-TPR$_\infty$+ & 323 & .879 & .24 & .930 & .950 & 37 & .616 & .521 & .674 & .786 \\ 
HMem-TPR$_\text{im}$ & 245 & .936 & .924 & .947 & .952 & \textbf{\underline{24}} & \textbf{\underline{.790}} & \textbf{.731} & \textbf{\underline{.831}} & \textbf{.886}\\\hline
\end{tabular}
\\ \vspace{.5cm}
\caption{Results on WordNet and Freebase benchmarks. 
Hyperparameters were tuned on the validation set for each model class M, with results recorded for both infinite and the best finite value of $\lambda$. Each model was evaluated on the test set using the same graph as in training, and also (M+) when extending the inference graph with all triplets from the validation set, without performing additional gradient descent on the validation triplets. Implicit binding models (M$_\text{im}$) forego explicit binding in favor of memory states that are directly learned for each entity. MR: Mean rank. MRR: Mean reciprocal rank|mean(1/rank). Hits@N: Proportion of test trials in which the rank of the test entity was less than or equal to N.  
$\dagger$: results from \citet{kaldec2017baselines}, who optimized hyperparameter settings for DistMult. } \label{table_main_results} }
\end{table*}


\section{Results} \label{section_results}

Models were evaluated using the benchmark datasets \textsc{WN18} (a subset of WordNet), \textsc{FB15K} (subset of Freebase), and the "challenge" dataset WN18RR, which removes reciprocal relation pairs from the training and test set of WN18, which can be solved by adopting a simple rule-based system \citep{dettmers2018convE}. 
We varied the binding method \{CConv, TPR\}, the value of optimization constant $\lambda$ \{$\infty$, 1, 2\}, and entity and relation embedding sizes. To illustrate the effect of each model component, we report results for both binding methods and best results from finite and nonfinite values of $\lambda$. If $\lambda = \infty$, the optimization step is the identity map, in which case the inference objective is to express the target binding as a linear combination of input bindings via memory weighting.

We report the standard evaluation metrics for the Link Prediction task, in which the model is queried on both the left and right sides, ranking candidates for each query. For instance, in the left-query $(\cdot, \boldsymbol{r}, \boldsymbol{e}_j)$ with true completion $(\boldsymbol{e}_T, \boldsymbol{r}, \boldsymbol{e}_j)$, each candidate entity $\boldsymbol{e}_c$ is scored as $\text{score}_\ell(\boldsymbol{e}_c|e_j,r)$ and the results are ranked. Mean Rank (MR) is the mean rank of the true candidate. The Mean Reciprocal Rank (MRR) is the average of $\frac{1}{rank(e_T)}$ for each true , a metric that is less sensitive to outliers. The Hits@N metric refers to the proportion of test triplets in which the true candidate appeared in the top $n$ entities, with Hits@1 denoting accuracy. In each case, all attested links (those found in the training, validation and test sets) are first filtered from the list of candidates. In Appendix 2, we also report the results from ablating conditioning of the weight matrix $\mathbb{W}_i$ on $\mathtt{M}_i$, instead using the global weight matrix $\mathbb{W}_\text{global}$.

A virtue of our model is that it can be freely augmented with additional graph triplets after training; hence, we also report results when including validation triplets in the graph used for inference (Model+). This introduces no bias in model selection, which is performed just on the training data. We also explicitly compare neighborhood aggregation to an embedding-based approach, the \emph{implicit binding} models (Model$_{im}$), in which memory states for each entity are learned directly as embeddings rather than being assembled from the entity neighborhood (\emph{explicit binding}). These remain binding models since they are treated identically to the explicit binding memories with respect to unbinding. For instance, the TPR$_{im}$ model predicts links by unbinding a predicted entity from the optimized memory state $\text{opt}(\mathtt{M}_i)$, where $\mathtt{M}_i$ is now a $m_E \times m_E$ tensor that is learned for each entity.

\begin{table}
{\footnotesize
\begin{tabular}{|l|c|c|c|c|c|} \hline
Model & MR & MRR & H@1 & H@3 & H@10 \\ \hline\hline
ComplEx$^\dagger$ & 5261 & .44 & .41 & .46 & .51  \\
ConvE & 5277 & .46 & .39 & .43 & .48 \\
ConvKB & 2554 & .248 & - & - & .525 \\ 
HypER & 5798 & .465 & .436 & .477 & .522 \\
\hline
CConv+ & \textbf{4609} & \textbf{.408} & .373 & \textbf{.427} & \textbf{.471}	\\ 
CConv+ &  7553 & .387 & .347 & .414 & .453	\\ 
CConv$_\text{im}$ & 4775 & .401 & \textbf{.381} & .408 & .437 \\ \hline
TPR+ & \textbf{\underline{2223}} & \textbf{\underline{.432}} & .384 & \textbf{\underline{.458}} & \textbf{\underline{.514}} \\ 
TPR+ & 3662 & .397 & .350 & .432 & .469  \\
TPR$_\text{im}$ & 3595 & .424 & \textbf{\underline{.393}} & .440 & .479 \\ \hline \hline
\end{tabular}}
\caption{Results on WN18RR. We compare with all models from Table \ref{table_main_results} where authors reported results on WN18RR. $\dagger$: results from \citep{dettmers2018convE}.}
\end{table}

Harmonic Memory models is state of the art for WN18 and Freebase \ref{table_main_results}, achieving especially noteworthy improvements in the Mean Rank metric. TPR binding outperforms circular convolution in every setting with quite low-dimensional embeddings (at most 80d for entities and 25d for relations). It is also competitive with recent models on WN18RR. As well, extending the graph with additional triplets after training yields improvements in all but one case (HMem-TPR on Freebase). The inclusion of the memory-completion module substantially improves performance on WordNet relative to M$_\infty$ on the more stringent evaluation metrics|leading for instance to an 8-point improvement in Hits@1 for HMem-TPR. The implicit binding models substantially outperform explicit binding on Freebase, a fact that is only true on aggregate, with important distinctions arising when entities with different neighborhood sizes are considered. We discuss this in Section \ref{section_saturation_effects}. 

\section{Generalizing to new entities} \label{section_generalizing}

To evaluate HMem's ability to generalize exclusively on the basis of aggregated neighborhoods, we introduce a new KBE task in which models make inferences about entities not seen in the training set. 
Consider the following scenario: a model is trained to complete a given knowledge base, but the knowledge base can be augmented in time not just with new facts about the current set of entities, but also with new entities. It would be desirable to perform inference over these new entities, without re-training the model, once partial information about these entities becomes available. 
Embedding-based models typically require that a representation for each entity be learned in the course of training. Hence, entities not encountered in the training set cannot be modelled. This creates a scalability problem: the knowledge base cannot be augmented with new entities without additional rounds of gradient descent. In contrast, our networks model entities by aggregating their links with other entities in the training graph, allowing entities not seen in training (and hence without a learned embedding) to be represented as memory states once information about these entities' neighborhoods becomes available. 

We built two datasets for knowledge base embedding with generalization (KBEGen) using WordNet (WN18) and Freebase (FB15K). First, a random selection of entities in each database (1500 for WN18, 1000 for FB15K) were randomly held-out, and all triplets not containing these entities were assigned to the training set. The number of entities held out was manually chosen to yield approximately the same training data size as the original datasets (WN18 = 141K, FB15K = 483K). Of the remaining triplets, we removed any for which both entities were part of the held-out set, and further split the remaining data into an \emph{observed subgraph} (2/3), a validation set (1/6), and a testing set (1/6). The observed subgraph was used to construct neighborhoods for each of the held-out entities, which were not trained with any further rounds of gradient descent and did not have trained entity embeddings. 

\textbf{Evaluation} We fit the model using the training set. In evaluation, the observed subgraph was used to construct a memory state for each entity in the held-out set using summation of entity-relation bindings in the observed subgraph for the held-out entity. For each test triplet, the memory state was probed using the query relation to rank the held-in entities as candidate neighbors for the modelled entity. 
\begin{table}[h!]
    \centering
    {\footnotesize
    \begin{tabular}{|l|l|l|l|l|l|} \hline
         & \textbf{heldout} & \textbf{train} & \textbf{valid} & \textbf{test} & \textbf{obs} \\ \hline
         WNGen & 1.5K & 141K & 1.7K & 1.7K & 6.8K \\
         FBGen & 1K & 496K & 15K & 15K & 62K \\ \hline
    \end{tabular} }
    \caption{Size of the KBC generalization dataset partitions WNGen and FBGen. \emph{heldout}: \# of held-out entities; \emph{obs}(erved): \# of triplets containing held-out entities that were retained for constructing the inference graph. 
    The number of entities held out for each dataset was manually chosen to yield approximately the same training data size as the original datasets (WN18 = 141,442, FB15K = 483,142). }
    \label{able_t}
\end{table}

\textbf{Results.} Performance on generating memory states for unseen entities (Table \ref{table_results_gen}) is far from ceiling but well above chance, with a more than 50\% accuracy (Hits@1) for the best-performing model on WordNet.\footnote{Performance with random initialization on WordNet is less than 1\%.} Notably, performance on WordNet improves dramatically from the addition of the validation subgraph during inference, leading to a nearly 10-point increase in accuracy for the best-performing model (CConv+). Improvements are smaller but reliable for FBGen.

\section{Scaling properties} \label{section_saturation_effects}

\begin{figure}[h!]
\centering
\includegraphics[scale=.45]{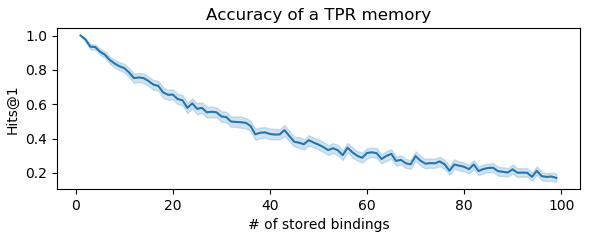}
\caption{\textbf{TPR memory}. Performance of a simulated TPR memory in retrieval of stored bindings of 100 100d "entity" vectors bound to twenty 20d "relation" vectors. For each simulation, we summed $n$ ($x$-axis) bindings of randomly generated relations to randomly generated entities. Decoding from the TPR degrades as the number of entity-relation bindings increases. } \label{fig_tpr_memory}
\end{figure}

As illustrated in Fig. \ref{fig_tpr_memory}, superposition memories are prone to increased decoding errors as the number of stored vectors increases. This is due to 
two factors: overlap between relation vectors even when these are linearly independent, and many-to-one nature of relation-entity bindings. Our model balances two competing priorities: (1) including as much information as is relevant for inference;  (2) reducing the number of stored entity-relation bindings, which tend to interfere with each other during retrieval.

\begin{figure}[h!]
    \centering
    \includegraphics[scale=.25]{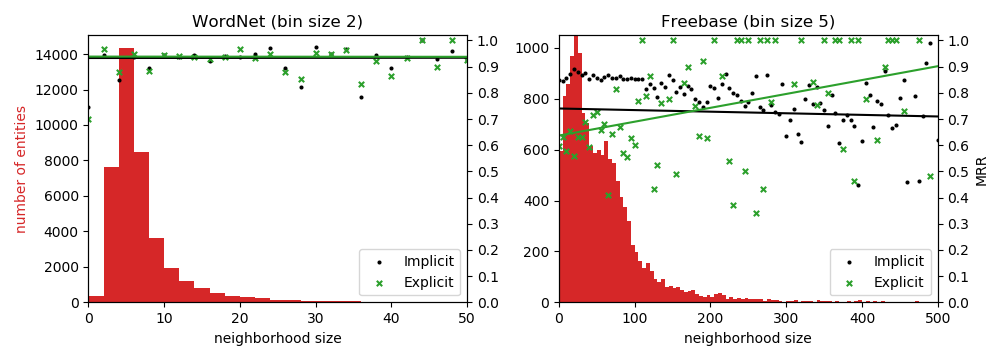}\\
    {\footnotesize 
\begin{tabular}{|l|l|l|l|l|l|l|} \hline
\textbf{Model} & \textbf{100} & \textbf{200} & \textbf{300} & \textbf{400} & \textbf{500} & \textbf{600} \\ \hline
Implicit & .862 & .816 & .793 & .702 & .741 & .617 \\ \hline
Explicit & .632 & .746 & .772 & .856 & .835 & .900 \\ \hline
\end{tabular} }
    \caption{Effect of entity neighborhood size on task MRR for the best implicit and explicit binding models in WordNet and Freebase, with neighborhood sizes binned at increments of 2 and 5 respectively. The line of best fit for neighborhood size against MRR is also plotted. \textbf{Table}: MRR for implicit and explicit binding models trained on \emph{FB15K} averaging over neighborhoods of different sizes (0 to 99 neighbors, 100-199 neighbors, etc.). }
    \label{fig_neighborhood_size_effect}
\end{figure}

The memory weighting module shoulders the burden of priority number (2), and is effective for this purpose (\ref{fig_neighborhood_size_effect}). We compared the embedding-based models with weighted explicit binding by considering performance as a function of the size of an entity's neighborhood (node degree). Explicit binding outperforms direct embedding in WordNet, though both methods are unaffected by neighborhood size. Large differences appear in Freebase, which has a much higher average node degree. Performance in the implicit model is highest in smaller neighborhoods and declines with node degree. With explicit memory construction, however, model performance is higher on nodes with large neighborhoods, peaking at an MRR of .9 for nodes with more than 500 neighbors. 

This appears counterintuitive given that  higher-degree nodes have more training in-
stances, which might yield higher-quality embeddings in the implicit models. We can explain the result. Consider the simplified scenario of a TPR memory trained to minimize the retrieval error for fixed entity/relation vectors with respect to the embedding $\mathtt{M}_\text{cat}$, we have:
\begin{align*}
    \mathcal{L} &= \mathbb{E}\left[ \left\lVert e_j - r_i \right\rVert\right] \\
            &=\sum_{i,j} p(\boldsymbol{r}_i,\boldsymbol{e}_j|\boldsymbol{e}_\text{cat}) \left\lVert e_j - r_i \cdot \mathtt{M}_\text{cat} \right\rVert^2
\end{align*}
$\mathtt{M}_\text{cat}$ is optimized\footnote{Up to a scaling factor, provided the components of the relation vectors $r_i$ are uncorrelated (i.e. the second moment $\mathbb{E}[{r_i {r_i}^\top}] \propto I$). This can be guaranteed by transformation (compare Appendix 1).} by
\begin{align*}
    \mathtt{M}_\text{cat} = \sum_{i,j} p(\boldsymbol{r}_i, \boldsymbol{e}_j | \boldsymbol{e}_\text{cat})\ r_i \otimes {e_j}
\end{align*}
i.e.\ an expectation-weighted superposition of pairwise vector outer products. In the KBC setting, the distribution  $p(\boldsymbol{r}_i, \boldsymbol{e}_j| \boldsymbol{e}_\text{cat})$ is uniform over all nonzero entity-relation pairs in the training set, meaning that entities with more neighbors have more nonzero terms in the solution (cf. fig. \ref{fig_tpr_memory}). The learned embedding is thus susceptible to increasing decoding error with increasing neighborhood size.

This result extends quite generally to a large class of models
|such as Rescal \cite{nickel2011rescal} and HolE \cite{nickel2016hole}|that we can formulate as binding models. For instance, the bilinear scoring function Rescal evaluates triplets by dotting left and right entity vectors with a relation-specific bilinear form $W_r \in \mathbb{R}^{d_e\times d_e}$: $\text{score}(\boldsymbol{e}_i,\mathtt{r},\boldsymbol{e}_j) = {e_i}^\top W_r e_j$. For given entity embeddings, the optimal relation embedding is $\hat{W}_r = \frac{1}{n} \sum_{i}^{n} e_{\ell,i} e_{r,i}^\top \equiv \frac{1}{n} \sum_{i}^{n} e_{\ell,i} \otimes e_{r,i}$ for all of the $n$ attested $\langle e_{\ell,i}, e_{r,i}\rangle$ edges involving $r$. This is a superposition of entity pairs bound by the tensor product where, to evaluate candidate links for the query $(\boldsymbol{e}_i,\boldsymbol{r}, \cdot)$, we first retrieve prototypical $\boldsymbol{r}$-neighbor for $\boldsymbol{e}_i$|$e_i \cdot W_r$, which is the weighted sum of all of the entities $\boldsymbol{e}_j$ that $\boldsymbol{e}_i$ occurred with|and then compare each candidate with this prototype. The greater the number of attested neighbors $\boldsymbol{e}_j$, the higher the anticipated retrieval error.

\begin{table}[h!]
\centering \scriptsize
\begin{tabular}{|l|l|c|c|c|c|c|} \hline
& Model & MR & MRR & H@1 & H@3 & H@10 \\ \hline\hline
WNGen 
& CConv & 2286 & .487 & .426 & .527 & .594	\\ 
& CConv+ & \textbf{1359} & \textbf{.592} & \textbf{.518} & \textbf{.647} & \textbf{.716}	\\ 
\cline{2-7}
& TPR$_\infty$ & 2127 & .435 & .373 & .476 & .540 \\ 
& TPR$_\infty$+ & 1507 & .514 & .448 & .565 & .624 \\ \hline \hline
{FBGen} 
& CConv$_\infty$ & 378 & .205 & .130 & .225 & .358 \\
& CConv$_\infty$+ & 373 & .207 & .131 & .251 & .361 \\ \cline{2-7}
& TPR$_\infty$ & 401 & .252 & \textbf{.173} & \textbf{.299} & \textbf{.439} \\
& TPR$_\infty$+ & \textbf{397} & \textbf{.263} & \textbf{.173} & \textbf{.299} & \textbf{.439} \\ \hline
\end{tabular}  \caption{Results on the KBE\textsc{Gen} task.} \label{table_results_gen}
\end{table}


The improvements from comparing explicit versus implicit binding can be attributed to the weighting module's judicious choice of information to include in a particular query. Notably, our results differ from those of \citet{schlichtkrull2017graphconv}, who found performance decreases with increasing node degree in a graph convolution-based model, indicating that our approach is promising to pursue in graphs with high mean node degree. 

\begin{figure}
    \centering
    \includegraphics[scale=.25]{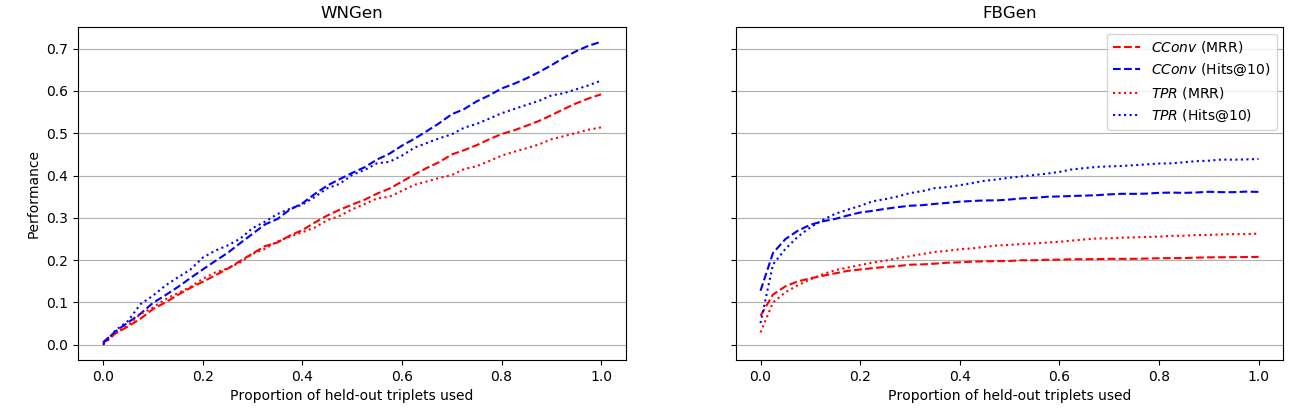}
    \caption{Performance on WNGen (left) and FBGen (right) test sets as a function of the proportion of observed graph triplets included in the inference graph. Each fraction of the full dataset (observed+valid) was used to construct the inference graph, the model then being evaluated on the test set. The model was not trained on any triplets from the observed subgraph or validation set.}
    \label{fig_mem_effects_gen}
\end{figure}

HMem with explicit binding also scales well with the open-world setting of evolving knowledge graphs. We note again that performance almost always increases on WordNet and Freebase when the inference graph is augmented with validation triplets on which the model was not trained. Fig. \ref{fig_mem_effects_gen} makes this point dramatically in the context of KBEGen. For entities held out from training, we gradually increased the proportion of the available inference graph|the entirety of which was held out in training|used to predict links from the test set. For both WNGen and FBGen, performance increases when new triplets are added, almost linearly in the case of WNGen. The concavity of the performance from added graph triplets in FBGen is likely not due to diminishing returns from the addition of information about particular entities, but rather to the fact that Freebase entities mostly have small neighborhoods, meaning that gains are felt mainly in the long tail. 

\section{Conclusion} \label{section_conclusion}

This article presents a neural model for knowledge base completion that is powerful enough to achieve state of the art results on large databases, and flexible enough to evolve with knowledge base content sans retraining. The approach complements existing neighborhood-aggregation techniques (e.g. graph convolution), with the advantage of interpretable mechanisms: vector binding and memory completion. The results indicate that the model operates well at scale and in an open-world setting.



\bibliographystyle{apalike}
\bibliography{research.bib}


\begin{table*}[h!]
\centering 
{\scriptsize
\hspace{-.65cm}
\begin{tabular}{|l|ccccc|ccccc|}
\multicolumn{1}{c}{\phantom{h}}& \multicolumn{5}{c}{WordNet} & \multicolumn{5}{c}{Freebase} \\ \hline
Model & MR & MRR & H@1 & H@3 & H@10 & MR & MRR & H@1 & H@3 & H@10 \\ \hline\hline
HMem-CConv  & 639 & .774 & .719 & .818 & .863 & 336 & .456 & .372 & .517 & .597 	\\ 
HMem-CConv+ & 475 & .787 & .730 & .880 & .952 & 361 & .442 & .362 & .500 & .575 \\ 
HMem-CConv$_\infty$  & 815 & .842 & .793 & .884 & .916 & 336 & .456 & .372 & .517 & .597 \\
HMem-CConv$_\infty$+ & 690 & .854 & .804 & .900 & .930 & 361 & .442 & .362 & .500 & .575 \\
\hline
HMem-TPR & 209 & .854 & .796 & .908 & .932 & 303 & .449 & .361 & .507 & .601 \\ 
HMem-TPR+ & 124 & .863 & .804 & .920 & .944 & 321 & .440 & .355 & .496 & .586 \\
HMem-TPR$_\infty$ & 154 & .866 & .784 & .923 & .950 & \textbf{36} & \textbf{.618} & \textbf{.523} & \textbf{.677} & \textbf{.786} \\
HMem-TPR$_\infty$+ & \textbf{110} & \textbf{.868} & \textbf{.796} & \textbf{.935} & \textbf{.963} & 37 & .616 & .521 & .674 & \textbf{.786} \\ \hline
\end{tabular}
\\ \vspace{.5cm}
\caption{Results of ablation of memory-conditioned weight matrix (see Appendix 2 text).} \label{table_ablation}} 
\end{table*}

\section{Appendix 1: Conditions on circularly correlated embeddings (Decorrelation transformation)}

As discussed in \citet{plate1994thesis}, a sufficient condition for circular correlation to approximately invert circular convolution is that the components of the vectors occurring in the memory are independently and identically distributed, with an expected norm of 1,\footnote{This can be guaranteed by setting the componentwise variance of the input vectors to $\frac{1}{d}$.} in which case the result the binding-unbinding sequence $x \star (x \circledast y)  = (1+\eta)y + \varepsilon$ where $\eta$ and $\varepsilon$ are zero-mean and approximately Gaussian noise terms \citep[pg 66]{plate1994thesis}. 

During learning, the entity and relations embeddings evidently depart from these strict conditions|as is desirable, since many of their latent features can and do covary. But to preserve the integrity of the decoding process, we enforce Plate's distributional constraints by applying a decorrelating transformation to the embeddings.\footnote{Early experiments confirmed that applying decorrelation improved performance with the CConv model.} 
At each step of training or inference, the entity and relation embeddings are concatenated, and the resulting array is centered by calculating the mean embedding $\mu$ and subtracting it from each embedding. Let $\boldsymbol{E}$ denote the $(|\mathcal{E}| + 2|\mathcal{R}|)\times d$ matrix of centered relation and entity embeddings. We calculate the empirical covariance matrix of the embeddings $\Sigma_\text{emp}$ and regularize it to produce an estimate $\hat{\Sigma}$. 
\begin{align*}
\Sigma_\text{emp} &= \frac{1}{|\mathcal{E}| + 2|\mathcal{R}|} \boldsymbol{E}^\top\boldsymbol{E} \\
\hat{\Sigma} &= (1-\alpha)\Sigma_\text{emp} + \alpha {I}
\end{align*}
$\alpha$ was set to .2. The regularized estimate of the covariance is then used to calculate the precision matrix $\hat{\Sigma}^{-1}$ for the centered embeddings. This defines a whitening transformation for any embedding $v$, obtained  by centering the vector and then post-multiplying it with the square root of the precision matrix, divided by $\sqrt{d}$ to ensure a variance of $\frac{1}{d}$ in every direction:
\begin{align*}
\hat{v} &= \frac{1}{\sqrt{d}}(v - \mu)\hat{\Sigma}^{-\frac{1}{2}}
\end{align*}
where $\mu$ is the average of all entity and relation vectors. The resulting distribution of transformed embeddings is approximately spherical with variance $\frac{1}{d}$ and an expected norm of $1$. When this transformation is applied to all vectors involved in binding and unbinding, Plate's conditions are met.

\section{Appendix 2: Ablation}

To evaluate the role of each model component, we ablated (1) the memory-completion operation, in which case the network's goal is to obtain held-out links as weighted sums of known links, and (2) conditioning the weight matrix in Eqn. \ref{eqn_max_mem_state} on the location of the input memory $\mathtt{M}_i$. Ablation (1) is implicit in setting the hyperparameter $\lambda = \infty$. We performed ablation (2) by keeping the weight matrix constant across all choices of $\mathtt{M}_i$, completing the memory using Eqn. \ref{eqn_max_mem_state} where $\mathbb{W}_i$ is set to $\mathbb{W}_\text{global}$.

\textbf{Results} The results of ablation (1) on the primary models are discussed in the main text. Ablation of the conditional weight matrix substantially performance with the convolution models, so that the best-performing models are those where $\mathbb{W}_i$ is recomputed for each $\mathtt{M}_i$ (Table \ref{table_ablation}). Ablated TPR models performed slightly better on Freebase, while on WordNet showed better performance on the Mean Rank metric, but were substantially outperformed by unablated TPR models on the key evaluation metrics, MRR and Hits@1, generally understood as the final arbiters of model performance. Interestingly, in contrast to the main results, where adding the validation graph to the TPR model improved performance on Freebase, doing so with the global weight matrix is marginally harmful.

\end{document}